# Dubito Ergo Sum:  Exploring AI Ethics

Viktor Dörfler
University of Strathclyde Business School
Glasgow, United Kingdom
viktor.dorfler@strath.ac.uk

Giles Cuthbert
Chartered Banker Institute
Edinburgh, United Kingdom
giles.cuthbert@charteredbanker.com

**Abstract**

*We paraphrase Descartes' famous dictum in the area of AI ethics where the "I doubt and therefore I am" is suggested as a necessary aspect of morality. Therefore AI, which cannot doubt itself, cannot possess moral agency.  Of course, this is not the end of the story.  We explore various aspects of the human mind that substantially differ from AI, which includes the sensory grounding of our knowing, the act of understanding, and the significance of being able to doubt ourselves.  The foundation of our argument is the discipline of ethics, one of the oldest and largest knowledge projects of human history, yet, we seem only to be beginning to get a grasp of it.  After a couple of thousand years of studying the ethics of humans, we (humans) arrived at a point where moral psychology suggests that our moral decisions are intuitive, and all the models from ethics become relevant only when we explain ourselves.  This recognition has a major impact on what and how we can do regarding AI ethics.  We do not offer a solution, we explore some ideas and leave the problem open, but we hope somewhat better understood than before our study.*

## 1. Introduction

In this conceptual paper we argue for the ethical approach in AI that suggests leaving most of the moral issues in the hands of humans.  This is not to say that we should not try to 'put' a moral perspective into AI, but that we also need to 'put' in the limitations.  The natural first step is that we need to understand the limitations, but the far trickier next one is how to get AI to identify its limitations, and request human assistance.  We do not intend here to get into the technical details of what can and needs to be done in AI, we remain at the level of philosophizing AI and the human mind in the context of ethics, thus problematizing AI ethics.  We do this from a distinct phenomenological position, within a moderate interpretivist paradigm (Dörfler, 2023b).

In this paper we do not provide a generic review of the AI literature, we only explain the basic concepts that we use in this paper here in the introduction.  Thus, for the purpose of this paper we use one of the oldest definitions of AI, back from the Dortmund days, according to which AI is loosely defined as machines that can accomplish tasks that humans would accomplish through thinking (e.g. Dörfler, 2020).

This definition does not say anything about AI accomplishing such tasks in a way that resembles human thinking; we do not see anything in this definition that implies that AI would think in the human sense of the word.  Importantly, AI as a field is not simply a study of the machines, it is as much the study of the human mind (for a more detailed description see e.g. Dörfler, 2022; Dörfler, 2023a).  Specifically in the area of decision-making we find Davenport's (2018, p. 44) description of AI as *"analytics on steroids"* particularly expressive and, consequently, AI does not make decisions but it can make our (human) decisions better informed.

Decision-making is an important aspect of using AI when it comes to ethics, and all (at least the vast majority of) our decisions have moral components.  In this paper we do not engage with particular application areas of AI, such as medical diagnosis (Davenport & Glaser, 2022; Davenport & Glover, 2018; Göndöcs & Dörfler, 2023), we locate our interest loosely in organizations (Csaszar & Steinberger, 2022; Davenport & Euchner, 2023; Davenport & Miller, 2022; Glikson & Woolley, 2020; Grodal et al., 2023; von Krogh, 2018; Leavitt et al., 2021; Lindebaum & Ashraf, 2021), in which concept we include business organizations, government institutions, as well as organizations, such as hospitals and universities, regardless of whether they are for profit or not.  We are conscious of the organizational learning aspects and implications of AI (Balasubramanian et al., 2020; Davenport & Ammanath, 2020; Davenport & Mittal, 2022, 2023; Göndöcs & Dörfler, 2022; Oliver et al., 2017; Pachidi et al., 2021; Raisch & Krakowski, 2021; Tschang & Almirall, 2021); although we cannot tackle these at a great depth here.

HICSS

In order to develop our argument eloquently captured paraphrasing Descartes, in what follows, we begin with a brief but systematic overview of the most common approaches and models in the domain of ethics. This is followed by a review of the AI ethics literature. Then we outline our philosophical position and methodological considerations, before getting to the points we want to make. Each of the next three sections provides a component of our conceptual analysis. First, we explore the sensory grounding of knowledge, showing how such sensory grounding, in a sense we attribute it to humans, is not possible in AI. Second, we suggest that AI lacks understanding, and we illustrate this with recent events in the AI landscape. Third, based on the literature, we argue that doubting oneself is an essential ingredient of morality, and we show that this one requirement also incorporates the previous two. In our final commentary we discuss what can be done, and offer three points, to make the best use of AI in morally acceptable ways and indicate areas of further research.

## 2. The Vast Landscape of Ethics

In this paper the term ethics is used to designate a branch of philosophy, the discipline that studies morality. In turn, morality refers to dealing with the issues of good and evil, right and wrong, responsibility, and such. In this sense, ethics is one of the oldest topics studied by humankind. In the Western tradition we see philosophy starting with discussing metaphysics followed by epistemology, which is closely followed by ethics, which Socrates has brought centerstage. This means that we have at least some two and half millennia of literature to cover, therefore our review will not be comprehensive, but our review is systematic in the sense that the main philosophical models are organized into categories (see Figure 1).

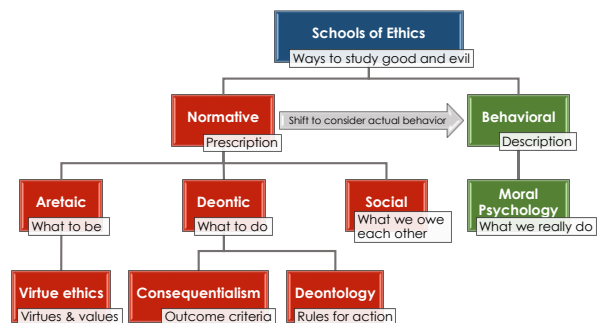

**Figure 1. Overview of Historical Schools of Ethics**

The normative schools attempt to prescribe what to be or what to do or not do; the behavioral school aims at describing what people actually do. The three dominant normative schools of ethics, are virtue ethics, rules-based ethics (or deontology), and consequentialism. Other important models in normative ethics include pragmatist, intuitionist, contractualist, and feminist ethics. While these schools are all centered on individuals, there are also less known social variants, proposed by a small number of philosophers.

While it was Socrates who initially turned the philosophers' attention to considerations of morality, particularly to counter the sophist approach to teach anyone how to win debates using tools of rhetoric, it was Aristotle (cca 330 BCE) who composed ethics into the first systematic model called *virtue ethics*. Aretaic ethics, of which virtue ethics is the dominant example, focuses on features or characteristics that are desirable in individuals if they are to be considered to meet high moral standards; in virtue ethics these characteristics are the virtues. Importantly, virtue ethics does not aim at providing answers or even tools to answering moral dilemmas, they describe the moral person. The main issues in virtue ethics is identifying what the relevant virtues are and how much of each is good, recognizing that virtues in excess may become vices (Hursthouse, 1999). Naturally, actions of people are considered morally correct if they embody the virtues required by the particular school; i.e. a moral dilemma can be answered by considering what a virtuous person would do in such situation. Therefore, there are no forbidden activities, so there is no tenet, for instance, not to kill, the dilemma is whether a virtuous person would kill.

After flourishing in Antiquity, virtue ethics was superseded by other schools, but it did have a revival in the 20-21 centuries, the leading figure being Alasdair MacIntyre (1998) and Elizabeth Anscombe (1958). MacIntyre (2007) also emphasized that we may need new models of virtue ethics for the modern world. Throughout history, and particularly during modernity, with its emphasis on scientific thinking and reductionism, attempts were made to derive all virtues from a single one, so can we say that all virtues are just consequences of e.g. courage or patience or, at least, construct a list of priority, i.e. virtues should be ordered by importance (or other organizing principle). Some recent philosophers, including Julia Annas (2011) and Anscombe (1958) argue that the virtues are all interconnected forming a complex system although the idea can be traced back to Aristotle (Gottlieb, 1984). They also discuss limitations of how actions of virtuous people may occasionally be wrong, which means that being virtuous is not a guarantee of right action. There are many other smaller schools of virtue ethics that we do not cover here. All schools of virtue ethics agree on two principles: they all attribute agency to humans, so that they can exercise their free will and choose to do the right thing, and they all consider humans rational and thus exercising rationality is part of being virtuous.

While areatic schools of ethics focus on what a virtuous person is like, deontic schools of ethics focus on what constitutes a morally right action. Deontology and consequentialism are both deontic (see Figure 1).

*Deontology*, in essence, offers sets of rules that we must follow to adhere to a high moral standard. Although there were a few earlier versions, the first comprehensive treatment of deontology comes from Immanuel Kant (1785, 1797). The central concept of Kant's deontology is the categorical imperative, according to which we should do what we would be happy to become a universal law. Another very important concept of Kant's deontology is the good will, meaning that if one has the right intentions, but things turn out badly due to things beyond one's control, that is acceptable. This served as basis for the development of a minority branch of deontology, which emphasizes intentions in contrast to consequences. While in these schools consequences do not matter, in most schools they do, only not as much as obligations and rights. A deontological model can incorporate one or both of two perspectives: agency and patiency. Agency links to obligations, i.e. the rules are based on what one is supposed to do. Patiency links to rights, i.e. the rules are based on how people can expect to be treated.

While for Kant rules are universally and infinitely valid, many approaches to deontology are more situational; sometimes it is OK to deviate from the rule for the sake of a better outcome. For instance, it is wrong to lie, but sometimes it may be the kind thing to do, and in some extreme situations (e.g. a monster asking you where your children are because it wants to eat them) it may be the only right thing to do. Of course, exceptions to rules can be formulated as rules, but only so far as we are able to foresee idiosyncratic situations. At first sight, deontology seems an excellent candidate for being considered in AI, as it is rule-based, and rules are easy to program. However, exceptions tend to be problematic in programming as much as in philosophy.

*Consequence ethics*, like deontology, is concerned with the right action, but not in terms of what we ought to do but in terms of the consequences of our actions. Consequences are typically expressed in the form of some utility or happiness at an individual or social level. Variants of consequentialism include hedonism, egoism, act consequentialism, and various forms of utilitarianism (e.g. Mill, 1861; Williams, 1993). There are two initial issues surrounding consequentialism: (1) Normally, we cannot know the consequences of our actions for sure at the time of taking the action. (2) Desirable consequences are often formulated as the best outcome for the greatest number of people, and thus the rights of minorities may be overlooked. Nevertheless, consequentialism seems to be, at least on a cursory look, to be well aligned with the idea of machine learning (ML) in artificial neural networks (ANN). Once we take further in-depth considerations, however, we find that there are other aspects of moral decisions that matter besides the consequences, even if we could be sure of those consequences; examples include the trolley problem and similar dilemmas (Foot, 1978; Williams, 1985).

There were a few individual approaches to ethics by leading thinkers of Enlightenment. These thinkers, including Georg Wilhelm Friedrich Hegel, Baruch Spinosa, David Hume, and Adam Smith emphasized the interconnectedness of people, leading to the view that other people are key to one's moral decisions. These approaches can be seen as a move from a fully individualist treatment of ethics towards a social perspective; for example, in a social variant of consequentialism we can consider what is good for society rather than the individual. An implication of these social approaches is the previously mentioned role of patiency in addition to agency in ethical considerations. A more contemporary example of a social view of ethics is Emmanuel Levinas (1961, 1991).

Besides these main normative schools, there are lesser-known ones, many of which can be, in some ways, considered "more human" than the previously introduced major schools. *Pragmatist ethics*, although not very prominently featured in the overall pragmatist philosophy, does not believe in the possibility of a single model that governs all moral decisions; ethics models can be useful, but different ones may be appropriate for different occasions (see e.g. James, 1899). According to *intuitionist ethics* we can know moral principles intuitively, as they are self-evident. These principles, typically duties (which makes intuitionism a form of deontology), are identified individually and therefore intuitionist ethics may be applicable in a greater number of situations than other approaches. For *contractualist ethics*, which is a form of deontology, justice makes an action right – although this is typically formulated in negative terms, i.e. avoiding wrongdoing and injustice. Finally, *feminist ethics* questions many assumptions of normative ethics models, which have all been conceptualized from a dominantly male, often patriarchal, perspective. This leads to rejecting the possibility of any absolute model in ethics (Gilligan, 2014), which means that both pragmatist and feminist ethics become pluralist and contextual.

We do not discuss the variants of ethical approaches in the Frankfurt School, as they are very fragmented and highly blended with political philosophy. What they agree about is similar to the starting point of feminist ethics, i.e. they all suggest that Western ethics is imbued with the exploitative values of the capitalist hegemony, making social justice impossible.

## 2.1. Moral Compasses

Moral psychology is about what people actually do rather than what they are supposed to do, therefore it can also be labelled as descriptive ethics in contrast with the normative schools (see Figure 1). The term moral psychology can be traced back to Anscombe (1958), who observed that the thinking in ethics should take into consideration what we learned studying psychology.

In one of the cornerstone works of moral psychology James Rest (1986) distinguishes four stages of moral decision making: moral sensitivity, moral judgment, moral motivation, and moral courage. A thorough treatment of moral psychology would require covering a significant amount of psychology literature, both conceptual and experimental, and making connection to the previously outlined normative approaches to ethics. Therefore, here we focus solely on one particularly important issue. Using Rest's stages, psychologists of ethics have established that people make moral judgments more or less exclusively using their intuition, and they only refer to the ethical models that they are familiar with when they need to justify their moral judgments to themselves or to others (cf Haidt, 2001). This is a very strong claim and moral psychology is largely in agreement about this point (Sonenshein, 2007). This is what we tried to capture with the notion of the *moral compass*.

However, we also must note that this does not mean that all ethicists, let alone scholars in the domain of AI ethics, subscribe to the dominant role of intuition in moral judgments; many emphasize deliberation with or without the use of normative models. There are also rare studies that try to synthesize the normative and descriptive approaches to ethics (Treviño, 1986). In our view, humans often arrive at moral judgments intuitively, just as in the case of any decision making, but this can also happen through sequential reasoning (Dörfler & Stierand, 2017). However, if consulting the models we are familiar with does not help to arrive at a moral judgment, we may revert to the use of intuiting as time may be pressing. A case can also be made that the *deliberation before action* is also only employed to justify the *intuitive judgment* already made. The debate is still ongoing, and we will not resolve it in this paper; for us it is important that intuitive judgments exist and both experimental and observational studies find significant use of intuition, particularly at a high level of mastery (Chase & Simon, 1973a; e.g. Chase & Simon, 1973b; Dörfler et al., 2009; Dreyfus, 2004; Dreyfus & Dreyfus, 1986; Gobet & Simon, 1996a, 1996b, 2000; e.g. Kreisler & Dreyfus, 2005). It is reasonable then to assume that moral judgments, like judgments more generally can be intuitive.

## 3. Ethics and AI

Ethical considerations in computer science are not new, and they get amplified in the world of AI. Norbert Wiener's (1960, p. 1358) formulation is still valid, perhaps more than ever:

*"If we use, to achieve our purposes, a mechanical agency with whose operation we cannot efficiently interfere once we have started it, because the action is so fast and irrevocable that we have not the data to intervene before the action is complete, then we had better be quite sure that the purpose put into the machine is the purpose which we really desire and not merely a colorful imitation of it."*

The literature on ethics in the scope of computers and wider digitalization, sometimes also referred to as digital ethics, largely applies normative models of ethics to the scope of the digital world (Anderson, 2011; Anderson & Anderson, 2011; Brey, 2000; Flanagan et al., 2008; Friedman et al., 2013; Moor, 1985; Vallor, 2016; van Wynsberghe, 2013). For example, Shannon Vallor (2016) seeks to adapt Aristotelian virtue ethics for the digital future.

The most important issue of AI ethics, as a scholarly discipline, is that it is almost completely conceptual. There are great explorations of applying a variety of normative ethics models in different digital/computerized/AI environments and theorizing or problematizing what the consequences would be – typically not leading to happy conclusions with the logical outcome that we may need a new normative model (Gunkel, 2017). Important problem areas in AI ethics, with reference to normative ethics, are agency, the roles emotions may play, levels of relativism, rationality, more specifically that there are different kinds of rationalities, the use of intuiting, as well as the relationship between a moral decision and action. For instance, rules-based ethics seems to be particularly suitable for computers, but whose rules to accept?

At the same time, AI vendors struggle with the ethical aspects of their products, and they keep looking at AI scholars and philosophers for help that they fail to provide. They look at the potential users of their products before designing a new product as well as after, and the opinions that they receive contradict each other and cannot be programmed. This problem, observed in the reality of AI vendors, is our starting point.

The most popular form of AI today, the deep neural networks (DNN) capable of deep learning (DL) cannot help. Much of the AI success today is ascribed to this form of AI, however, those successes must be looked at in context: the type of problems DL was applied to. In principle, DNN is simply an ANN with more than one hidden layer, and DL is a really efficient form of ML in a DNN, but the principles are not substantially different: ML needs a large number of learning examples and then it replicates the statistical frequencies of the outcomes with reference to the input variables (LeCun et al., 2015; Marcus, 2018; Schmidhuber, 2015). AlphaGo (Silver et al., 2016) needed some 300 billion games to get trained and deliver the extraordinary performance of beating Lee Sedol (Hassabis, 2017). So, if we even had a database of moral decisions, what number of learning examples would be needed for DL to produce the statistical frequencies? What variables would need to be considered? How many types of moral decisions are there? The list of questions could continue, each and every one of them would be sufficient to conclude that this is not the way to successfully deal with AI ethics. Furthermore, there is evidence that if the training data is biased, the ANN will amplify these biases.

Essentially, we need to understand that the problem of AI ethics is not an implementation problem. It is not about having the right conceptual construct that we need to operationalize, we are struggling with fundamental problems of ethics. Therefore, we need to step back and look into making progress in the field of ethics in which both humans and AI are present. So, what can be done? In order to figure this out, we look into three aspects of how AI is different from the human mind.

## 4. Methodological Considerations

We take a phenomenological approach in this conceptual study, framed within the broadly considered paradigm of Critical Interpretivism (Dörfler, 2023b). this means that our approach is moderate subjectivist, and we practice bracketing through transpersonal reflexivity to arrive at insights (Dörfler & Stierand, 2021). We do not adopt a theoretical lens, as any lens limits what can be seen, instead we adopt the approach known as phenomenon-driven theorizing, which allows us to approach the phenomenon at hand with an open mind and allow theorizing to take us in various directions (Fisher et al., 2021; Langley, 2021; Ployhart & Bartunek, 2019). The particular type of theorizing we undertake is called problematizing, as the purpose of it is not to provide a solution but to arrive at an improved understanding of what the problem is. To problematize AI ethics, we make use of everyday well-known phenomena, as it is often done in Gestalt psychology (Köhler, 1959; Rock & Palmer, 1990), and attempt to explain these employing abductive reasoning (Sætre & Van de Ven, 2022).

## 5. Indwelling

Sensing, i.e. sensory input is indispensable aspect of all knowing and is necessarily employed on par with intellect (Bas et al., 2022). Michael Polányi (1966b, p. 15), the renowned philosopher of knowledge, argues that the body is the ultimate instrument of all external knowledge. Antonio Strati (2007) wonders why the important role of the body is neglected and often ignored although it is the body that enables both intellectual reasoning and sensory-based knowledge. Sensing is not a unitary construct. Based on Burton (2009, p. 37) Dörfler and Bas (2020) consider, besides perception based on the five primary senses, also visceral sensations (e.g. hunger), affective sensations (e.g. love), as well as mental sensations (e.g. pride). With this expanded view of sensing, we can easily conclude that everything we know we come to know through sensing (Bas et al., 2022; de Rond et al., 2019; Strati, 2007). Furthermore, based on Polányi, we introduce the notion of indwelling, through the use of which the idea of sensing can be extended to abstract domains, such as mathematics or astrophysics, or microbiology: these are all abstract in the sense that we cannot get in touch with the subject of inquiry through our body, but the phenomenon is essentially the same. We must emphasize that we do not argue for the empirical over the rational, we suggest considering indwelling in addition to, rather than instead of, reasoning.

Why is this so important? Of course, people sense and there are various mechanical, electronic, etc. sensors that we can connect to computers. Surely a camera with the right set of calculations is more reliable than a human eye… However, two harsh critiques of AI, Hubert Dreyfus (1992) and John Searle (1994) both consider the computers' lack of sensory capacity in producing knowledge as one of the main reasons that computers cannot think and that the "strong AI" paradigm is impossible. A full treatment of this issue would entail exploring the issue of primary and secondary qualities, originally introduced by René Descartes (1637) and then elaborated by John Locke (1690) and later George Berkeley (1878), and deriving the notion of "felt sense" from these (Dörfler, 2023b). Therefore, we do not intend to engage in a generic debate about the possibilities of human-level (or nearly human-level) AI; we simply acknowledge the significance of sensory grounding in ethics in the light of the four phases of moral decision-making identified in moral psychology. As before, we emphasize that we are not arguing that sensing should replace reasoning but that sensing and reasoning are both essential.

## 6. Understanding

There is no complete agreement in cognitive psychology or the philosophy of mind regarding the precise definition of the concept of understanding. Russell Ackoff (1989) locates it between knowledge and wisdom. We know that tacit knowledge plays a crucial role in understanding, as Polányi (1966a, p. 7) suggests:

*"While tacit knowledge can be possessed by itself, explicit knowledge must rely on being tacitly understood and applied. Hence all knowledge is either tacit or rooted in tacit knowledge. A wholly explicit knowledge is unthinkable."*

The significance of understanding in ethics is perhaps obvious: if we are to make moral decisions, we need to be able to understand that our actions have consequences (consequentialism), even if we do not exactly know what the consequences are, as need to understand the rules that we are meant to follow (deontology), we need to understand the implications of specific virtues on our actions (virtue ethics).

Debates on whether AI has or can have understanding are as old as AI. In 1957 Herbert Simon predicted four things AI was supposed to achieve within ten years (Simon & Newell, 1958, pp. 7-8). The only one that has been achieved is that a computer has beaten the best chess player in the world, but that only happened in 1997. Simon has also asserted:

*"I believe that in our time computers will be able to perform any cognitive task that a person can perform. I believe that computers already can read, think, learn, create…" (Simon, 1977, p. 6)*

Some recent events, however, may rejuvenate these discussions. A chess robot broke a seven-year-old-boy's finger (Henley, 2022). The Bing AI chatbot called a CNN reporter "rude and disrespectful", presumably for asking too many questions (Kelly, 2023), and declared love to a NYT journalist and tried to convince him that he did not love his wife, but the chatbot (Roose, 2023). There are also numerous examples of factual and logical mistakes made by ChatGPT-4. However, there was a particularly instructive story that happened in February 2023. Kellin Pelrine, an American amateur Go player (ranked one level below the top in amateurs) beat the top Go computer 14 out of 15 games (Waters, 2023). What makes this point significant, is that all 14 times it was the same trick. If the computer had any level of understanding of the game, it would have identified the same trap being set the second time, let alone being tricked 14 times the same way. However, these examples only showcase that AI does not possess understanding right now, but not that it cannot.

Of course, there are many examples of generative AI (and other forms of AI) delivering incredible performance, the question is whether AI can understand and if not now, can AI ever understand (Chomsky et al., 2023). We believe that the examples conclusively prove that AI does not understand. We also believe that AI is not designed to think, but to mimic some of the outcomes of thinking. Clearly, there are other opinions, in the end it all boils down to whether we accept the computational model of the mind. The view of understanding has a significant impact on the view of AI ethics, as moral judgments presume understanding.

## 7. Doubting

Finally, the third pillar of our argument is the capacity to doubt. When we paraphrase René Descartes, we accentuate that he was not a Sceptic (one of the four Hellenistic philosophical traditions, of which David Hume was a late follower). Descartes was trying to combat the sceptic dictum that we can doubt everything by attempting to find solid ground in those things that we can be really sure about. To this end, he adopted the Sceptic armament and applied it to everything that he could think of, in a systematic doubt. In doing so, he behaved like a Sceptic, demonstrating how we can doubt anything that we think we know. We cannot be sure that it is an object in front of us, we cannot even be sure of our own bodies, it can all be an illusion, an evil demon's doing who hijacked our minds. And then, in a master-stroke, he turns the argument upside down, and concludes that if one can doubt anything and everything, then there is one thing that one can be sure of, and it is that there *is* something that can doubt. As doubting is a form of thinking, Descartes (1637, p. 27) formulates his famous dictum: *"Cogito ergo sum"* (I think and therefore I am). However, it is more precise if we limit the term to doubting, and thus 'Dubito ergo sum' (I doubt and therefore I am). What we are suggesting here, is that being moral entails the capacity to doubt, we need to be able to doubt our actions, to doubt ourselves (de Crescenzo, 1992; e.g. Spiegelberg, 1947).

So, what does it mean to be able to doubt, particularly in the context of ethics? It entails sensing our decision situation and understanding it, trying to do the right thing but not being able to figure out what our actions may lead to, reflecting and not being sure even of our motivations. Just think of Hamlet's painful struggle whether he should avenge his father. Doubt incorporates some of the most complex issues of the human mind, including sensing and understanding, and it may well be indispensable for our moral decisions, for our moral development, perhaps the central component of the moral mind. There is no consensus about this point, the significance of doubt is our own observation in the realm of AI ethics. Doubt also has an interesting implication regarding certainty: there is a significant body of literature on uncertainty in entrepreneurship, strategy, and decision making, since Knight (1921, 1923) suggested that our default condition is uncertainty, in which alternatives and their respective probabilities are not known. To cope with uncertainty, we make social contracts, perhaps we can think of doubt in a similar vein. Doubt scarcely appears in the AI literature (see Shklovski & Némethy, 2023 for a rare example). Importantly for AI ethics, if doubt seem to be essential for our moral judgments, what are the implications of doubt-less AI? Importantly, while we developed our argument from a phenomenological perspective, the concept of doubting can be significant for AI in any philosophical position.

## 8. Final commentary

In several ways, AI ethics is a weird concept. It can cover moral considerations of making AI, ethical aspects of using AI, potential consequences of the tasks we assign to AI, and so forth (Asaro, 2006; Floridi & Sanders, 2002). When we build AI, if we hand over some of our decisions to it, we need to put in something that takes care of those aspects that constitute the moral dimensions of our decisions. How can we do that? What should it be? We know, for instance, that if trained on biased data, AI may amplify those biases.

We will not pretend to have figured out how to build a moral engine for AI. However, we now perhaps understand a little better what needs to be considered for such an attempt. This is incredibly timely, it was as we were writing the first version of this paper that we found out that Microsoft has sacked all its AI ethics team just as they were getting ChatGPT into Bing and soon possibly into many other products. This means one thing for us: ethical questions of AI are difficult, complex, ignite heated debates – and it is paramount that we get them right.

It looks like there is no easy way that would let us 'program' ethics into AI, or let it learn it through ML/DL. The reason is that we do not have a proverbial perfect moral entity whose moral characteristics or actions we could use as starting points. Even if we could find such an entity, we would struggle to identify a sensible number of learning examples – and we could not even start figuring this out, as we have no idea how many types of moral decisions exist. There is also no large model that could be used by generative AI.

What does this leave us with? Well, there are a few things we can pin down: (1) The main point of moral psychology was that we make at least some of our moral judgments intuitively and only use ethics models to explain them. (2) Our moral judgments are rooted in indwelling (typically sensory perception) and we need to understand the decision situation as well as the possible consequences of our actions. This does not mean that we can know the consequences, but we can think up scenarios. (3) The capacity to doubt seems an indispensable part of moral decisions. None of this, however, suggests to us that AI ethics is a futile area.

What we are trying to figure out is what AI can help us with in terms of moral decisions and to understand the way forward. There are two immediate things that we believe AI can do for us right now: (1) AI can provide us with useful input for our process of self-doubt (ex ante or ex post), as we are deliberating our moral decisions by identifying potentially relevant patterns in available ethics models. This can help in two ways, it can reduce the struggle of self-doubt and can help us explain our moral judgments. (2) AI can scan the context for emerging information and patterns, feeding this back to us so that we can course-correct quickly. Humans rely on their 'felt sense', like babies calling for their parents when they need a change of diapers: they feel uncomfortable. As AI lacks 'felt sense', we need to figure out how to provide external pointers when AI needs to involve a human in the process.

Considering the previous discussion we now make a leap and suggest something that does not trivially follow from what has been said. AI is an amplifier. It does not make us smarter, it amplifies what we have, and if we are stupid, it will amplify that as well (Dörfler, 2022). We have seen e.g. how AI can amplify biases. However, the leap is the following: we suggest that we do not actually have AI ethics problems. This is the reason that we went all the way back and took a journey in time for 2.5 millennia. We do not have *AI ethics problems*, we have *ethics problems*. AI amplifies them.